\documentclass[preprint,12pt]{elsarticle}
\usepackage{amssymb}
\usepackage{amsmath}
\usepackage{url}
\usepackage{multirow}
\usepackage{amsfonts}
\usepackage{bbm, dsfont}
\usepackage{caption}
\usepackage{lscape}
\usepackage{subcaption}
\usepackage{longtable}
\usepackage{tabularx, booktabs}
\usepackage{natbib}
\usepackage[table,xcdraw]{xcolor}

\journal{Informatics in Medicine Unlocked}
\begin{document}
\begin{frontmatter}
\title{EnLVAM: Enhanced Left Ventricle Linear Measurements Utilizing Anatomical Motion Mode\tnoteref{tn1}}
\tnotetext[tn1]{Visual Intelligence publications are financially supported by the Research Council of Norway, through its Center for Research-based Innovation funding scheme (grant no. 309439), and Consortium Partners. The work was further partially funded by RCN FRIPRO grant no. 315029.}
\author[1]{Durgesh K. Singh\corref{cor1}}
\ead{durgesh.singh@uit.no}
\cortext[cor1]{Corresponding author}
\author[4]{Ahc\`ene Boubekki}
\ead{ahcene.boubekki@ptb.de}
\author[2]{Qing  Cao}
\ead{qing.cao@gehealthcare.com}
\author[3]{Svein Arne Aase}
\ead{sveinarne.aase@gehealthcare.com}
\author[1]{Robert Jenssen}
\ead{robert.jenssen@uit.no}
\author[1]{Michael Kampffmeyer}
\ead{michael.c.kampffmeyer@uit.no}
\address[1]{Department of Physics and Technology, UiT The Arctic University of Norway, Tromsø, Norway}
\address[2]{GE Healthcare, Wuxi, China}
\address[3]{GE Vingmed Ultrasound, Horten, Norway}
\address[4]{Machine Learning and Uncertainty Group, Physikalisch-Technische Bundesanstalt, Berlin, Germany}

%% Abstract
\begin{abstract}
Linear measurements of the left ventricle (LV) in the Parasternal Long Axis (PLAX) view, using two-dimensional Brightness-mode (B-Mode) echocardiography videos, are essential for evaluating cardiac conditions. These measurements involve placing four or six landmarks in a straight line called virtual scanline~(SL) perpendicular to the LV long axis at or just below the level of the mitral valve leaflet tips. However, manual placement of these landmarks is time-consuming and prone to errors. While current B-mode-based deep learning methods aim to automate this process by directly predicting key points, they often result in misaligned landmarks, leading to inaccurate measurements and the over- or underestimation of LV dimensions. In this study, a novel framework is proposed for enhancing left ventricle (LV) linear measurements by incorporating straight-line constraints. A landmark detector is trained on Anatomical Motion Mode (AMM) images, which represent various motion patterns of LV structures. These images are computationally derived in real-time from B-mode echocardiography video and selected based on a specific SL choice. The predicted AMM coordinates from the landmark detector are then transformed back to the B-mode space. By training the neural network in the AMM image space, misalignment issues are addressed by the proposed framework, with overestimation and underestimation errors being reduced through the enforcement of the straight-line constraint. Experiments demonstrate that the proposed framework achieves improved accuracy compared to B-mode methods using the same detector. Furthermore, the framework is versatile and can be applied across various neural network architectures.
Additionally, the semi-automatic design of our approach incorporates a human-in-the-loop component, where the human is only required to place an SL, a simpler task compared to full landmark placement. This design ensures that the method adheres to the flexibility required by medical guidelines in the real-world, while maintaining high measurement accuracy.
\end{abstract}


\end{frontmatter}
\section{Introduction}\label{sec:introduction}

Echocardiography~(Echo) is a widely used cardiac imaging technique for screening and diagnosing heart diseases, valued for its affordability, non-invasive nature, and ability to capture real-time heart images~\citep{Maus2022}. Accurate and reproducible measurements on echocardiography images are critical for precise clinical diagnosis. LV linear measurements are among the most frequently used parameters for evaluating LV cavity size and mass, which are significant indicators of LV hypertrophy \citep{Foppa2005}. LV linear measurements are performed by capturing frames in the parasternal long axis (PLAX) view using either 2D Brightness-mode~(B-mode) or 2D Echo guided Motion-mode~(M-mode) and three LV linear measurements are performed: interventricular septum (IVS), LV internal diameter (LVID), and LV posterior wall (LVPW)~\citep{solomon2019153}. 

\begin{figure*}[!t]
	\includegraphics[scale=0.2]{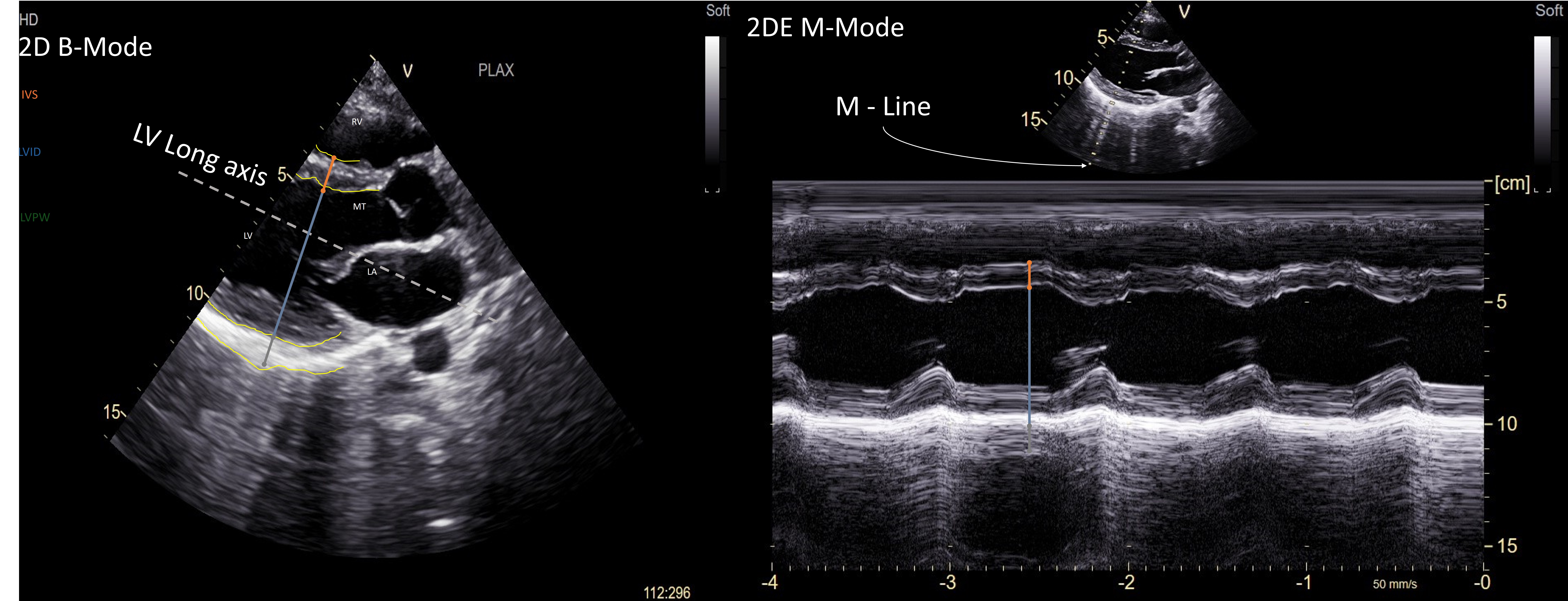}
	\caption{LV linear measurement in Parasternal Long Axis view~(PLAX). LV: Left Ventricle; RV: Right Ventricle; MT: Mitral valve leaflet tips; RA: Right aritrum; IVS: Interventricular septum; LVID: LV internal diameter;  LVPW: LV posterior wall. Myocardial wall boundaries are shown in yellow.}
	\label{fig:annotation}
\end{figure*}

The process begins typically by identifying two key frames, known as ES~(end-systole) and ED~(end-diastole), representing ES and ED phases of the cardiac cycle. B-mode measurements are performed by placing four or six calipers on the image grid. Alternatively, in M-Mode, the sonographer first places a scan line~(M-line) at the LV level on the B-mode image, and then an M-mode image is generated which traces the LV structures in time along the M-line throughout different cardiac cycles. The calipers are placed along the M-line in the resulting M-mode image~(example provided in Figure~\ref{fig:annotation}). 

ASE guidelines recommend performing LV linear measurements in the B-mode, where measurements should be performed at the basal level i.e.  \textit{``the level at or just below the mitral valve leaflet tip''} and perpendicular to the LV long axis~\citep{Lang2015,Mitchell2019}. However, differences in image quality and subjective interpretations of the mitral valve leaflet tips can result in inaccurate manual measurements. Additionally, the guidelines suggest that all measurements should move slightly toward the LV apex to just beyond the septal bulge, where the septum is abnormally thickened (\citet{Mitchell2019}). This seems to conflict with the mitral valve leaflet tips level in normal cases.

\begin{figure}[!t]
	\centering
	\includegraphics[scale=.15]{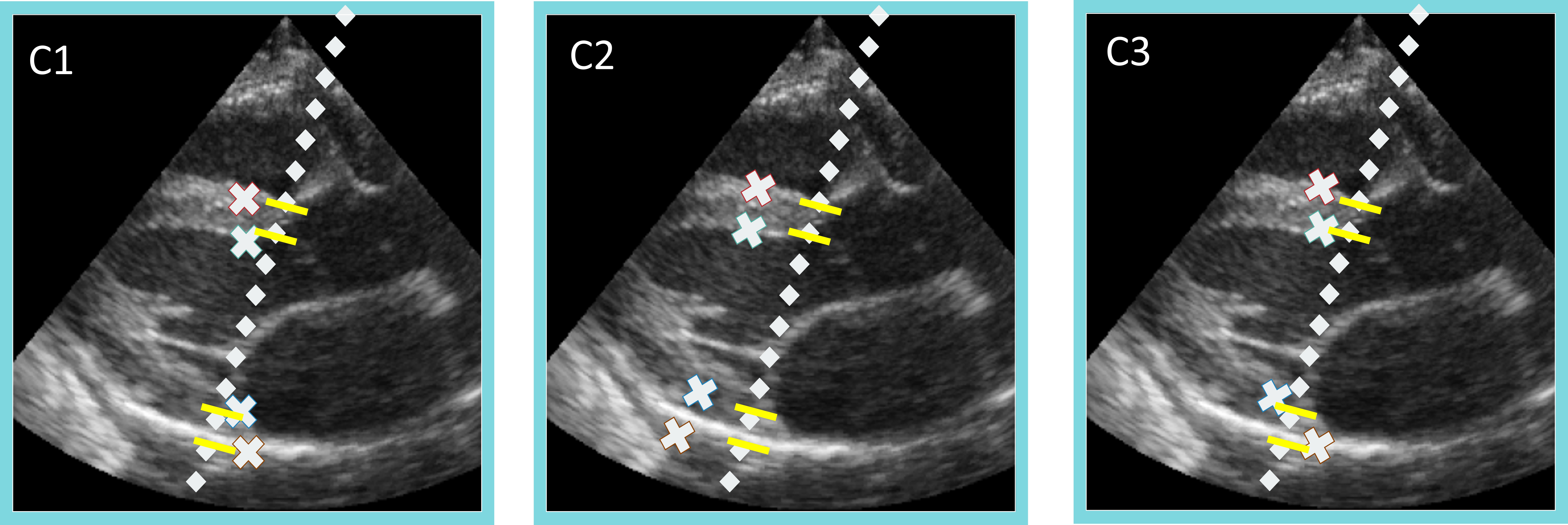}
	\caption{ 
		Cases~(C1, C2, C3) show incorrect predictions~(Section~\ref{sec:introduction}) from existing automatic B-mode approaches. The line~($\diamond$) marks the optimal virtual scanline~(SL) position and orientation according to ASE guidelines~\citep{Mitchell2019}, while the calipers~(-) indicate landmark locations for LV linear measurement.}
	\label{fig:predictions_fully_automated}
\end{figure}

Recently, deep learning-based methods have been developed to enable fast and reproducible LV linear measurements. These models utilize deep landmark detectors as the base model, training a neural network to predict heatmaps from ES/ED keyframes in B-mode Echo videos. These heatmaps indicate the likelihood of a landmark's presence at specific locations within the image grid. However, the scarcity of high-quality training data often results in imperfect heatmaps, leading to inaccurate landmark predictions unsuitable for precise LV linear measurements. Figure~\ref{fig:predictions_fully_automated} illustrates the limitations of prior automatic LV model: predicted landmarks may not align perpendicularly to the LV long axis (Figure~\ref{fig:predictions_fully_automated}, C1), may shift from the basal level (Figure~\ref{fig:predictions_fully_automated}, C2), or, more commonly, may randomly deviate along the myocardial wall (Figure~\ref{fig:predictions_fully_automated}, C3). These inaccuracies in automatic B-mode approaches introduce risks of over- or under-estimation when deriving linear measurements from predicted LV landmarks. Moreover, B-mode automatic landmark predictors lack the flexibility required for broader clinical applicability, particularly in  scenarios such as septal bulge cases. While ASE guidelines emphasize measurements at the basal level, a study by \citet{Chetrit2019} offers an alternative perspective, highlighting the importance of performing measurements at the LV mid-ventricular level for various clinical purposes. However, existing automatic approaches are not adaptable to accurately predict landmarks at the mid-ventricular level, limiting their utility in addressing diverse diagnostic requirements.

% This variability leads to significant inter- and intra-observer discrepancies. Furthermore,
% measurements should be taken around the mitral valve leaflet tip in cases where there is a septal bulge, indicating an abnormally thickened septum. Finally, identifying the correct M-line for generating M-mode images can be difficult and time-consuming for clinicians. As a result, guidelines from the American Society of Echocardiography (ASE)~\citep{Lang2015,Mitchell2019} emphasize the importance of careful execution of LV linear measurements in the B-mode to mitigate the errors regarding the incorrect placement of LV landmarks.

 In this study, we explore the benefits of using the Anatomical Motion Mode (AMM) imaging technique~\citep{Carerj2003} for performing LV measurements, addressing the limitations of previous automatic approaches. AMM images are similar to 2DE M-mode images but differ in their method of generation. Unlike 2DE M-mode, which relies on the clinician’s selection of an M-line during the examination, AMM images are computationally generated from pre-recorded B-mode echocardiography videos. This process involves tracing LV structures over time along a user-defined virtual scan line~(SL) that is oriented perpendicular to the LV long axis (see Figure \ref{fig:amm_annotation}).
 Clinicians use AMM images to incorporate the time component, enabling them to better visualize LV motion and dynamics over cardiac cycles, which can help correct or refine B-mode measurements for improved accuracy.

This study proposes training a base landmark detector on real-time computed AMM images by transforming B-mode coordinate annotations into AMM space. This adaptability is especially valuable in challenging clinical scenarios, such as cases with septal bulge, where precise SL placement above the mitral valve leaflet tips is essential for the accurate measurements. By addressing these varying situations, the AMM-based approach broadens its applicability in diverse clinical contexts, making it a powerful tool for LV measurements. Experimental results show that the AMM-trained landmark detector provides improved accuracy by resolving the misalignment issue present in B-mode approaches when the optimal SL is provided during training. We summarize the contribution of our work as follows:

\begin{figure}[!t]
	\centering
	\includegraphics[scale=.15]{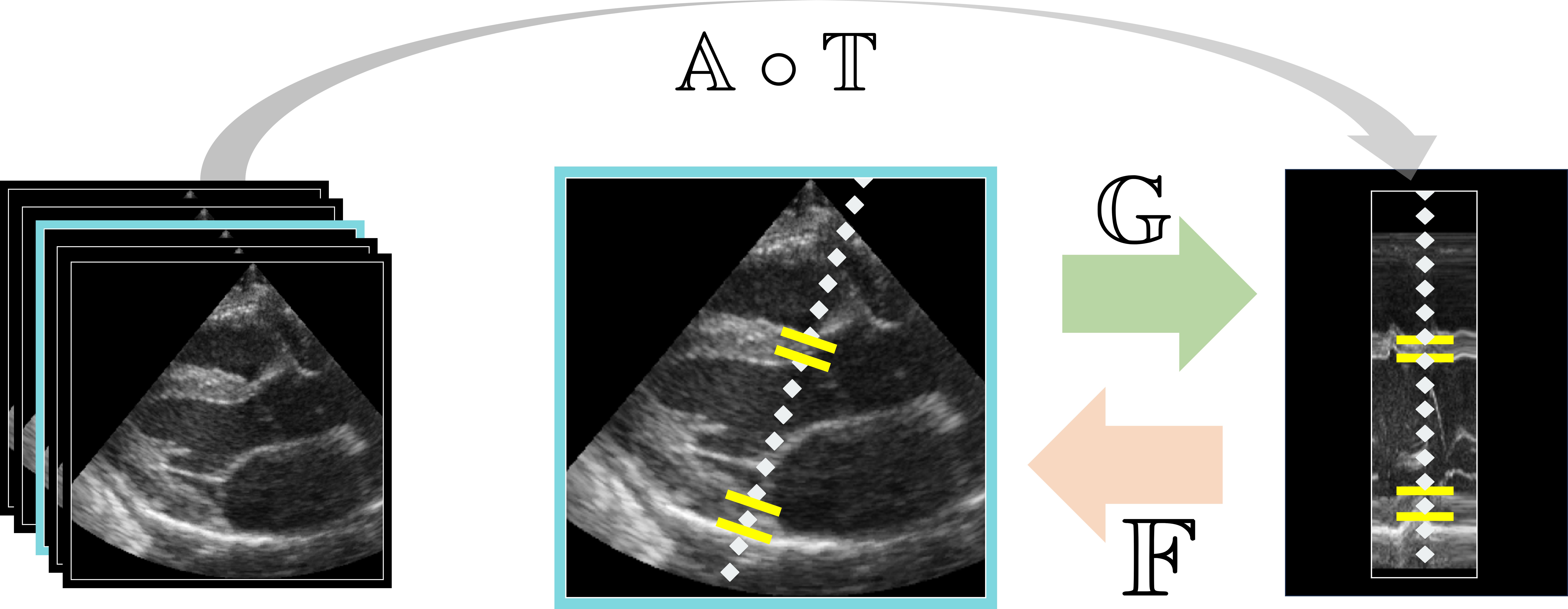}
	\caption{Measurements using B-mode and AMM images. The AMM image (right), resembling real M-mode images, is computed by selecting an anchor frame (\(\mathbbm{A}\)) and tracing (\(\mathbbm{T}\)) LV structures over time. Measurements in yellow are transformed to AMM space using \(\mathbbm{G}\), with \(\mathbbm{F}\) as vica-versa.}
	\label{fig:amm_annotation}
\end{figure}

\begin{itemize}
	\item[-] We address the shortcomings of B-mode  approaches by utilizing the advantages from the AMM imaging, representing the structures movement on a virtual SL over time and generate AMM images without requiring additional images input from users.
    
	\item[-] The semi-automatic design of our model facilitates the incorporation of different user preferences and allows flexible LV linear measurements at different SL positions during inference, such as the basal or mid-ventricular levels, all without the need for additional training labels.
	
\end{itemize}

The paper is structured as follows. Section~\ref{sec:background} covers landmark prediction in medical imaging and discusses related prior LV landmark detection methods, placing this work in their context. In Section~\ref{sec:method}, the problem is formally outlined, and a detailed description of the proposed solution, i.e., AMM LV semi automatic linear measurements, is provided. Sections~\ref{sec:experiments} to \ref{sec:results} cover data processing, experimental setups, baseline comparisons and discuss results comparing the performance of the two training modes. Section~\ref{sec:sdr} presents a detailed comparative analysis of AMM and B-mode training through the success detection rate curve. This analysis highlights the key differences between the two approaches and evaluates the clinical utility of each method, providing valuable insights into their respective strengths and applications. 
Section~\ref{sec:visualization} presents visualizations that demonstrate improved LV linear measurements achieved with AMM across various clinical scenarios. Lastly, Section~\ref{sec:conclusion} summarizes the findings, situates our contribution within a broader context, and outlines avenues for future research.

\section{Background}\label{sec:background}
In this section, related works for landmark detection in the medical imaging field are first discussed. Subsequently, the literature on relevant LV landmark detection works is reviewed to place our work in a broader scope.

\subsection{Landmark Detection in Medical Images}\label{sec:ldmi}
Heatmap-based landmark detection approaches are common in medical images~\citep{Chen2022} and can be categorized into two main approaches: (i) Regression-based and (ii) Classification-based~\citep{Zhou2014}. Regression-based landmark detection utilizes a distance-based loss to directly regress heatmap values based on ground-truth annotations. In these methods, a convolutional neural network (CNN) processes the 2D image to generate heatmaps, which indicate the probability of a landmark being present at specific locations within the image grid. The heatmaps are then processed by a coordinate extraction scheme to output the final location of the landmark. Classification-based methods predict the probability that a landmark is present in predefined grid locations within the image. This is achieved by dividing the input image into grids or candidate locations, where the regression-based method generates a 2D heatmap of confidence scores, indicating the presence of a landmark at each grid location. A thresholding scheme is then applied to determine the presence or absence of the landmark at each specific location. 
Classification-based approaches struggle with quantization errors and are heavily dependent on input grid resolution. Higher-resolution inputs improve accuracy but require more computational resources and training data. Regression-based approach can achieve subpixel-level accuracy for coordinate extraction, with label smoothing further enhancing their performance~\citep{Szegedy2016}. However, uniform label smoothing can lead to imperfect heatmap prediction, and thereby less effective for identifying landmarks at fuzzy boundaries in medical images. To address this, several methods have been developed to improve boundary-aware landmark placement during heatmap regression~\citep{Feng2018, Zhang2020}.
The choice between regression-based and classification-based approaches in medical imaging depends on the specific task requirements. Regression-based methods perform well with effective label smoothing, while classification-based methods benefit from high input grid resolution. However, both approaches face challenges in accurately placing landmarks for precise LV measurements due to imperfectly predicted heatmaps and do not account for user preferences for the SL.

\subsection{LV Landmark Detection}
In this section, deep LV landmark detection approaches are discussed. \citet{Sofka2017} performed landmark predictions from heatmaps using a center of mass layer and long short-term memory \citep{Hochreiter1997}. \citet{Gilbert2019} performed landmark predictions for IVS, LVID, and LVPW simultaneously, using the differential spatial to numerical (DSNT) transform to extract landmark locations \citep{Nibali2018}. \citet{Lin2021} used a shared encoder along with a detection and tracking head for landmark prediction and tracking in a single B-mode cine loop. \citet{Howard2021UIC} utilized a High-Resolution Network (HRNet) for LV landmark detection. \citet{Jafari2022} developed a framework using Bayesian U-Net \citep{Ronneberger2015} to automate the keyframe selection task and landmark detection in echo videos. \citet{Duffy2022} used atrous convolutions \citep{Chen2018} to learn the dimensions of the left ventricle and trained a residual network-based model in parallel for disease classification. \citet{Wan2023} utilized a ResNet-50 \citep{He2016} as the encoder part of a U-Net and placed a CoordConv layer \citep{Liu2018} after the network input for improved performance.

Although these works have effectively utilized heatmap regression for landmark detection, they face challenges with precise landmark placement due to label smoothing. To improve landmark prediction, \citet{Mokhtari2023} used a classification-based approach, employing a hierarchical graph neural network-based decoder to accurately place landmarks for LV linear measurement. This approach includes a multi-resolution landmark detection framework and hierarchical supervision using a multilevel loss. However, achieving precise landmark locations remains challenging due to the resolution of the input images.
Recently, \citet{Jeong2024} utilized 2DE M-mode imaging to acquire high temporal resolution images from scanners in their research. They created an AI-enhanced system that includes a view classifier, a segmentation network, and an auto-measurement network to automate measurements on M-mode images. However, the training is confined to M-mode images obtained from specific transducer orientation at the LV level, limiting the model's adaptability to different SL preferences perpendicular to the LV long axis and thus potentially impacting the accuracy and flexibility of measurements. 
In contrast, our approach overcomes these limitations by computing AMM images in real-time from B-mode echocardiograms and training the landmark detector on the same.  Moreover, the trained model can accommodate different orientations of SL by directly computing AMM images along a user-preferred SL$_{test}$ during the inference, providing greater flexibility for the LV linear measurement without any additional training supervision, when compared to the B-mode approach. A detailed description of our method is provided in the next section.

\section{Method}\label{sec:method}

\subsection{Problem Formulation}\label{subsec:problemformulation}
In this section, a formal description of LV linear measurement task from a learning perspective is presented. Our objective is to utilize a neural network model to predict the landmarks on the ES/ED keyframe in an echocardiography video. To formally define this task, the following notation is used: Let the training dataset contain $M$ single-cycle videos, denoted as $\{F_m: m \in M \}$, each comprised of 2D B-mode 1-channel frames resized to $256 \times 256$ pixels. Each training sample $m$ in the dataset includes an anchor frame $K_{m}$ which is either ES or ED frame, located at the index $\mathbbm{A}(K_m)$. Furthermore, there are $N = 4$ coordinate annotations, provided in a straight line representing LV linear measurements. The goal of the trained landmark detector is to take an ES/ED keyframe as input and perform linear measurements by identifying the corresponding LV landmarks.

\subsection{Proposed AMM Solution}\label{subsec:ammsolution}
\begin{figure*}[t]
\includegraphics[scale=0.03]{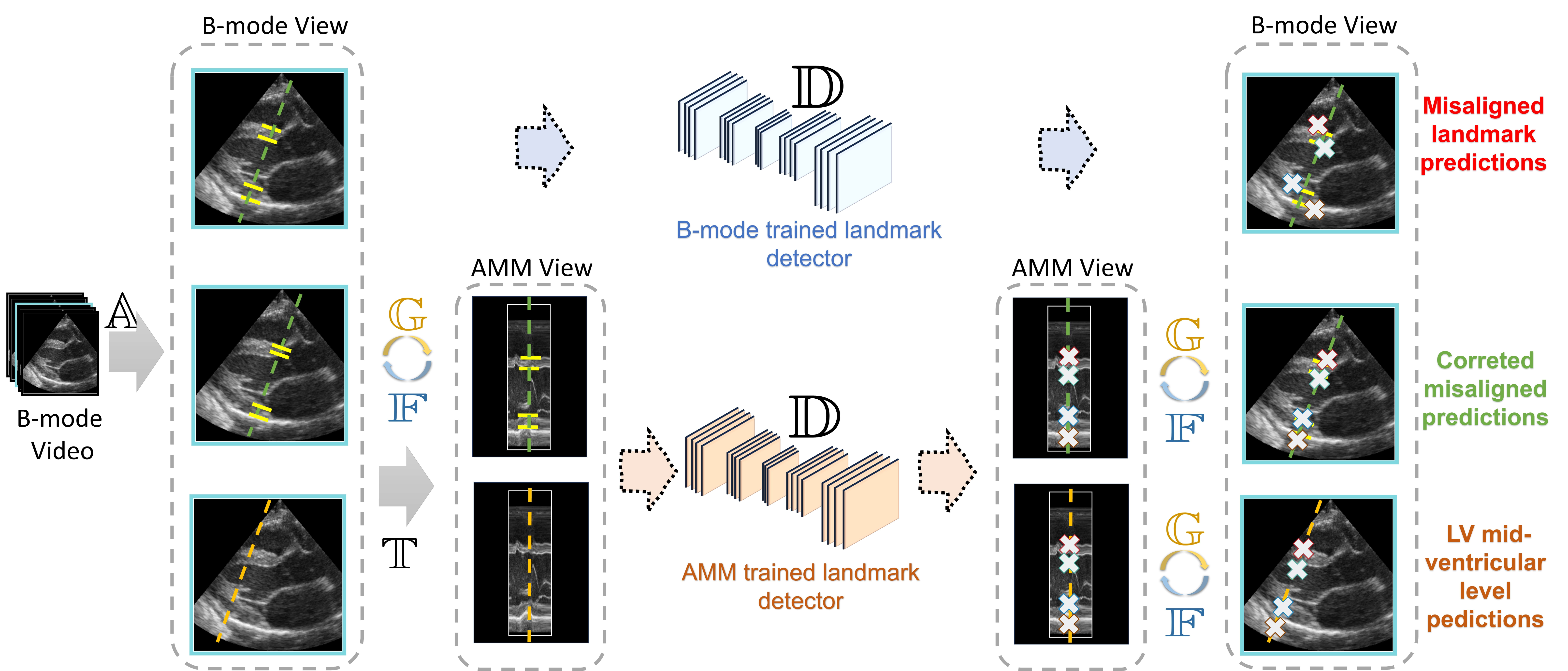}
	\caption{Figure describing the our AMM approach when compared to B-mode training; 
 $\mathbbm{A}$: Select the keyframe from Echo video, $\mathbbm{T}$: Tracing the LV in time, $\mathbbm{G}$: Convert B-mode landmarks to AMM, $\mathbbm{F}$: Convert AMM landmarks to B-mode, $\mathbbm{D}$: Base landmark detector}
	\label{fig:approach}
\end{figure*}

In this section, our approach, aimed at addressing the issue of shifted landmarks in prior works, is now described and a schematic diagram of our proposed solution is depicted in Figure~\ref{fig:approach}. Given input $m$, a clip $f_m$ of $W$ frames is extracted from the B-mode echocardiography video $F_m$, with the anchor frame $K_m$ centered. Additionally, $N=4$ B-mode coordinate annotations are provided for $K_m$ along a user-preferred SL. Our approach can be outlined as follows:

\textbf{Creation of AMM image from B-mode:}
An AMM image is first generated from \( f_m \) by tracing the LV in time with a bilinear interpolation function \( \mathbbm{T} \) along \( V \) equidistant points on the SL, which are obtained from co-linear ground-truths denoted as the set \( \{ p^v_m = (px_{m}^v, py_{m}^v): \forall v \in V \} \).

\begin{equation}
	\mathbbm{T} : (p_m^v, f_m^w) \mapsto A_m=[a_m^{vw}]_{(V \times W)}, \quad \forall v \in V, w \in W
\end{equation}

Here, \(A_m\) represents the resulting 1-channel AMM image, and \(\mathbbm{T}\) operates on the \(V\) equidistant points for each frame \(f^w_m \in f_m\) to generate the interpolated values \(a_m^{vw}\) on a grid of size \(V \times W\).  We further clarify that in \(A_m\) the SL is positioned at the center x-location.

\textbf{Base model training in AMM space}:
To train the base model $\mathbbm{D}$ for detecting LV landmarks using AMM images, and to facilitate supervised training, the B-mode ground-truth landmarks are first transformed as follows:
\begin{equation}
	\mathbbm{F}(C^{i}_m) : (x^{i}_{m}, y^{i}_{m}, W) \mapsto \left(\frac{W}{2}, \textsc{NN}(y^{i}_{m},\{py^{v}_m\})\right) , \quad \forall i \in N, v \in V
\end{equation}
Where, \(\mathbbm{F}\) maps a B-mode coordinate $C^{i}_m=\left(x^{i}_{m}, y^{i}_{m}\right)$ to AMM by assigning the x-coordinate location to the center of the anchor frame i.e $W/2$, and the y-coordinate is replaced by the nearest neighbor ($\textsc{NN}$) from the set \(\{py^{v}_m\}\).
Additionally, an \(N\)-channel heatmap \(H_m\) is created using AMM ground truth, by placing a unit Gaussian heatmap at the location \(\mathbbm{F}(C^{i}_m)\). We employ heatmap regression-based loss functions, which were introduced and utilized during the training process of the B-mode baseline models.

\textbf{Inference from the AMM trained base model}: Given an AMM-trained base model ($\mathbbm{D}$), our approach requires a user to provide a preferred SL, i.e~SL$_{test}$ during the inference. An AMM input $A_{test}$ from the test clip $f_{test}$ and SL$_{test}$ is first created, as before, by interpolating along $V$ equidistant points on SL$_{test}$. Next, AMM landmarks from $\mathbbm{D}$ are predicted as follows :

\begin{equation}
	\left(\hat{x}^{i}_{test}, \hat{y}^{i}_{test}\right) = \left(\frac{W}{2}, \sum_{v \in V} v \cdot \left[\text{S}(\hat{H^{i}}_{test})\right]_{\frac{W}{2},v}\right)\quad \forall i \in N,
\end{equation}

Where the softmax function \text{S} is used to normalize the predicted AMM heatmaps and the predicted y-coordinate ($\hat{y}^{i}_{test}$) for landmark $i$ is computed using the soft-argmax strategy from the heatmap $\hat{H^{i}}_{test}$ along the vertical axis at the center x-location~($W/2$) in $A_{test}$.
Here, we emphasize that, unlike its B-mode counterpart, the benefit of AMM training ensures the landmark locations are predicted onto the given SL$_{test}$. Finally,  the predicted AMM landmarks are transformed back to the B-mode anchor frame $K_m$ by utilizing the  same $V$ equidistant points   on the SL$_{test}$:

\begin{multline}
	\mathbbm{G}(\left(\hat{x}^i_{test},\hat{y}^i_{test}\right)) : (\left(\hat{x}^{i}_{test}, \hat{y}^{i}_{test}\right), \{p^v_{test}\}) \mapsto p^{v_i^{*}}_{test},\\ \quad\text{and}\quad v_i^{*}=arg_\textsc{NN}(\hat{y}^{i}_{test},\{py^{v}_{test}\}) \quad \forall i \in N, v \in V,
	\label{eq:bmode_argnn}
\end{multline}

Where $\mathbbm{G}$ transforms AMM predicted coordinates i.e. $\left(\hat{x}^i_{test},\hat{y}^i_{test}\right)$ to B-mode  coordinates i.e. $p^{v_i^{*}}_{test}$ onto the SL$_{test}$ for landmark $i$, and mapping $arg_\textsc{NN}$ returns the index $v_i^{*}$ of $p^{v_i^{*}}_{test}$~ having the nearest y-coordinate location with $\hat{y}^{i}_{test}$.

\section{Experiments}\label{sec:experiments}
In this section, we cover data processing, experimental setups, and baseline comparisons. 

\subsection{Training Dataset}
Our training and evaluation dataset is an extended version of the one introduced by \citet{Gilbert2019}. It comprises 493 echocardiography videos from 306 patients, each captured in the PLAX view and containing a single cardiac cycle, resulting in a total of 986 images. The dataset is divided with patient-level separation into 70\% for training (further split approximately into 90\% training and 10\% validation for 10-fold cross-validation) and 30\% for testing.

\subsection{Baselines}\label{subsec:baselines}
We compare our AMM-based approach, which constrains landmark predictions along a straight line, to the B-mode training approach. For a quantitative comparison, we evaluate the B-mode results using AMM-generated images with SL, derived from the ground truth annotations in the training dataset. This evaluation is referred to as AMM$_{\text{SL}}$. Additionally, we compare the B-mode results by projecting the predicted LV coordinates onto the scanline, aligning them along a straight line corresponding to the given SL. This projection forms the B-mode$_{\text{SL}}$ baseline. Further analysis of the performance differences between these two training modes is presented in Section~\ref{sec:results}, with success detection rate analysis discussed in Section~\ref{sec:sdr}.

\subsection{Model Training}
To accommodate training on a single GPU with limited memory (12GB), all models were trained for 60 epochs using a batch size of 2 with 8 gradient accumulation steps. The Adam optimizer was employed alongside a multistep learning rate scheduler, which reduced the learning rate by a factor of 0.1 at epochs 20 and 40.

\subsection{Evaluation Metrics}\label{subsec:evalmetrics}
We utilized three metrics to evaluate the performance of our neural network:

\begin{enumerate}
	\item Mean Absolute Error (MAE):
	\\
	\begin{align}
		\text{MAE} &= \frac{1}{N} \sum_{ {i,j \in N} \atop {j=i+1} }\left| \hat{C}^{ij} - C^{ij} \right|
	\end{align}
	MAE computes the average difference between the predicted lengths $\hat{C}^{ij}$ and the actual lengths $C^{ij}$ for the corresponding coordinates $\{C^{i},C^{j}\}$.
	The MAE metric offers a smoother representation of the distance-based error on the validation set, which makes it valuable for assessing the convergence of training.
	\item Mean Absolute Percent Error~(MAPE):
	\\
	\begin{align}
		\label{eq:ade_apde}
		\text{MAPE} &= \frac{1}{N}\sum_{ {i,j \in N} \atop {j=i+1} }\frac{|\hat{C}^{ij} - C^{ij}|}{C^{ij}}
	\end{align}
	
	The MAPE metric is a variation of the MAE metric, calculating the average percentage difference from the ground truth. It is particularly useful for assessing prediction accuracy relative to the data's scale, as it presents the error as the percentage. However, the MAPE metric often witnesses in high variance, especially with small LV structures i.e. IVS and LVPW, where even slight prediction errors can lead to disproportionately large percentage errors. 
	
	\item \textit{Coordinate Error~(CE):}
	\\
	\begin{equation}
		CE =  \frac{1}{N}\sum_{ {i \in N} }\| C^{i} - \hat{C}^{i} \|_2 
	\end{equation}
	The coordinate error~(CE) quantifies the Euclidean distance between the ground truth coordinates $C^{i}$ and the predicted coordinates $\hat{C}^{i}$ for each landmark. This metric provides a reliable measure of accuracy, taking into account any inaccuracies in the individual shifting of landmark coordinates. 
\end{enumerate}

Since MAE and MAPE may lack sensitivity to variations in which the predicted landmarks are shifted along the myocardial wall, as illustrated in Figure~\ref{fig:predictions_fully_automated}, we therefore consider CE as a robust metric for indicating shifted landmark prediction.

\section{Results}\label{sec:results}

In this section, we present the results of the AMM$_{\text{SL}}$ training compared to its B-Mode baseline, for different landmark detection architectures.
\\

\textbf{MAE}: The most significant improvements in MAE are observed in the LVID dimension across all three base models by approximately 0.15 centimeter~(cm). These reductions highlight the strong impact of AMM${_\text{SL}}$ on improving LVID measurements. The next most significant improvement occurs in the IVS dimension, where AMM${_\text{SL}}$ reduces the MAE by 0.08 cm for \citet{Duffy2022}, by 0.03 cm for \citet{Gilbert2019}, and by 0.06 cm for \citet{Wan2023}. Finally, improvements in the LVPW dimension, although notable, are the least pronounced: for \citet{Duffy2022}, the improvement is 0.06 cm; for \citet{Gilbert2019}, it improves by 0.05 cm; and for \citet{Wan2023}, the reduction is 0.09 cm. Overall, AMM${_\text{SL}}$ shows consistent improvements across all measurements, with the most significant changes in LVID, followed by IVS, and LVPW.
\\

\textbf{CE}: The CE metric measures the accuracy of predicted coordinates in relation to the actual positions, and
by aligning the predicted B-Mode landmarks onto a straight line SL, i.e. B-Mode$_{\text{SL}}$ , improves the accuracy of coordinates compared to B-Mode, reducing overestimation and underestimation errors. Comparing B-Mode$_{\text{SL}}$ to AMM based training, i.e. AMM$_{\text{SL}}$, we further observe the following improvements in mean errors: For \citet{Duffy2022}, the improvement is 0.17 cm, for \citet{Gilbert2019}, the improvement is 0.10 cm, and for \citet{Wan2023}, the improvement is 0.11 cm. These results highlight the effectiveness of AMM$_{\text{SL}}$ in further reducing errors, ensuring more accurate and reliable cardiac measurements. These reductions highlight the effectiveness of restricting the predicted coordinates onto SL using AMM space achieve superior accuracy across all base models.

\begin{longtable}{p{0.2\textwidth} p{0.2\textwidth} p{0.2\textwidth} p{0.2\textwidth}}
\caption{Table comparing B-mode and AMM training modes across different base models using MAE, MAPE, and CE for various LV dimensions;\quad B-mode$_{\text{SL}}$ represents B-mode predicted landmarks projected onto the SL, while AMM$_{\text{SL}}$ computes metrics with the SL$_{\text{test}}=$SL;\quad Statistically significant results are highlighted in bold (p-value $\leq$ 0.05);\quad cm: centimeter, pp: percentage points.}
\label{tab:lvam-results}\\
\hline
\textbf{\begin{tabular}[c]{@{}l@{}}Training \\ Mode\end{tabular}} & \textbf{IVS} & \textbf{LVID} & \textbf{LVPW} \\ \hline
\endfirsthead
\multicolumn{4}{c}%
{{\bfseries Table \thetable\ continued from previous page}} \\
\endhead
\hline
\endfoot
\endlastfoot
                     & \multicolumn{1}{l}{}   & \multicolumn{1}{l}{}   & \multicolumn{1}{l}{}   \\
\multicolumn{4}{c}{\cellcolor[HTML]{9B9B9B}{\color[HTML]{FFFFFF} \textbf{MAE (in cm)}}}         \\
                     & \multicolumn{1}{l}{}   & \multicolumn{1}{l}{}   & \multicolumn{1}{l}{}   \\
                     & \multicolumn{3}{c}{\citet{Duffy2022}}                                    \\
B-Mode               & 0.26$\pm$0.02          & 0.44$\pm$0.03          & 0.27$\pm$0.02          \\
B-mode$_{\text{SL}}$ & 0.27$\pm$0.01          & 0.46$\pm$0.05          & 0.28$\pm$0.04          \\
AMM$_{\text{SL}}$    & \textbf{0.18$\pm$0.01} & \textbf{0.29$\pm$0.01} & \textbf{0.21$\pm$0.01} \\
                     & \multicolumn{1}{l}{}   & \multicolumn{1}{l}{}   & \multicolumn{1}{l}{}   \\
                     & \multicolumn{3}{c}{\citet{Gilbert2019}}                                  \\
B-Mode               & 0.20$\pm$0.01          & 0.38$\pm$0.01          & 0.23$\pm$0.01          \\
B-mode$_{\text{SL}}$ & 0.19$\pm$0.02          & 0.36$\pm$0.03          & 0.21$\pm$0.01          \\
AMM$_{\text{SL}}$    & \textbf{0.17$\pm$0.01} & \textbf{0.23$\pm$0.01} & \textbf{0.18$\pm$0.01} \\
                     & \multicolumn{1}{l}{}   & \multicolumn{1}{l}{}   & \multicolumn{1}{l}{}   \\
                     & \multicolumn{3}{c}{\citet{Wan2023}}                                      \\
B-Mode               & 0.23$\pm$0.03          & 0.42$\pm$0.04          & 0.28$\pm$0.04          \\
B-mode$_{\text{SL}}$ & 0.22$\pm$0.02          & 0.40$\pm$0.03          & 0.25$\pm$0.04          \\
AMM$_{\text{SL}}$    & \textbf{0.17$\pm$0.01} & \textbf{0.27$\pm$0.01} & \textbf{0.19$\pm$0.01} \\
                     & \multicolumn{1}{l}{}   & \multicolumn{1}{l}{}   & \multicolumn{1}{l}{}   \\
                     & \multicolumn{1}{l}{}   & \multicolumn{1}{l}{}   & \multicolumn{1}{l}{}   \\
                     & \multicolumn{1}{l}{}   & \multicolumn{1}{l}{}   & \multicolumn{1}{l}{}   \\
\multicolumn{4}{c}{\cellcolor[HTML]{9B9B9B}{\color[HTML]{FFFFFF} \textbf{CE (in cm)}}}          \\
                     & \multicolumn{1}{l}{}   & \multicolumn{1}{l}{}   & \multicolumn{1}{l}{}   \\
                     & \multicolumn{3}{c}{\citet{Duffy2022}}                                    \\
B-Mode               & 0.78$\pm$0.08          & 0.82$\pm$0.10          & 1.01$\pm$0.10          \\
B-mode$_{\text{SL}}$ & 0.36$\pm$0.03          & 0.31$\pm$0.02          & 0.42$\pm$0.08          \\
AMM$_{\text{SL}}$    & \textbf{0.19$\pm$0.00} & \textbf{0.22$\pm$0.01} & \textbf{0.26$\pm$0.02} \\
                     & \multicolumn{1}{l}{}   & \multicolumn{1}{l}{}   & \multicolumn{1}{l}{}   \\
                     & \multicolumn{3}{c}{\citet{Gilbert2019}}                                  \\
B-Mode               & 0.74$\pm$0.04          & 0.76$\pm$0.05          & 1.10$\pm$0.06          \\
B-mode$_{\text{SL}}$ & 0.27$\pm$0.03          & 0.24$\pm$0.02          & 0.28$\pm$0.02          \\
AMM$_{\text{SL}}$    & \textbf{0.17$\pm$0.02} & \textbf{0.18$\pm$0.01} & \textbf{0.20$\pm$0.01} \\
                     & \multicolumn{1}{l}{}   & \multicolumn{1}{l}{}   & \multicolumn{1}{l}{}   \\
\multicolumn{1}{c}{} & \multicolumn{3}{c}{\citet{Wan2023}}                                      \\
B-Mode               & 0.73$\pm$0.04          & 0.75$\pm$0.01          & 1.07$\pm$0.06          \\
B-mode$_{\text{SL}}$ & 0.31$\pm$0.03          & 0.26$\pm$0.02          & 0.35$\pm$0.06          \\
AMM$_{\text{SL}}$    & \textbf{0.20$\pm$0.01} & \textbf{0.21$\pm$0.01} & \textbf{0.24$\pm$0.02} \\
                     & \multicolumn{1}{l}{}   & \multicolumn{1}{l}{}   & \multicolumn{1}{l}{}   \\
                     & \multicolumn{1}{l}{}   & \multicolumn{1}{l}{}   & \multicolumn{1}{l}{}   \\
                     & \multicolumn{1}{l}{}   & \multicolumn{1}{l}{}   & \multicolumn{1}{l}{}   \\
\multicolumn{4}{c}{\cellcolor[HTML]{9B9B9B}{\color[HTML]{FFFFFF} \textbf{MAPE (in percentage points)}}}          \\
                     & \multicolumn{1}{l}{}   & \multicolumn{1}{l}{}   & \multicolumn{1}{l}{}   \\
                     & \multicolumn{3}{c}{\citet{Duffy2022}}                                    \\
B-Mode               & 0.27$\pm$0.02          & 0.11$\pm$0.01          & 0.25$\pm$0.02          \\
B-mode$_{\text{SL}}$ & 0.26$\pm$0.01          & 0.11$\pm$0.01          & 0.25$\pm$0.04          \\
AMM$_{\text{SL}}$    & \textbf{0.18$\pm$0.00} & \textbf{0.07$\pm$0.00} & \textbf{0.20$\pm$0.01} \\
                     & \multicolumn{1}{l}{}   & \multicolumn{1}{l}{}   & \multicolumn{1}{l}{}   \\
                     & \multicolumn{3}{c}{\citet{Gilbert2019}}                                  \\
B-Mode               & 0.21$\pm$0.01          & 0.10$\pm$0.00          & 0.22$\pm$0.01          \\
B-mode$_{\text{SL}}$ & 0.19$\pm$0.01          & 0.09$\pm$0.01          & 0.19$\pm$0.01          \\
AMM$_{\text{SL}}$    & \textbf{0.17$\pm$0.01} & \textbf{0.06$\pm$0.00} & \textbf{0.17$\pm$0.01} \\
                     & \multicolumn{1}{l}{}   & \multicolumn{1}{l}{}   & \multicolumn{1}{l}{}   \\
                     & \multicolumn{3}{c}{\citet{Wan2023}}                                      \\
B-Mode               & 0.24$\pm$0.03          & 0.11$\pm$0.01          & 0.28$\pm$0.04          \\
B-mode$_{\text{SL}}$ & 0.22$\pm$0.02          & 0.10$\pm$0.01          & 0.24$\pm$0.03          \\
AMM$_{\text{SL}}$    & \textbf{0.17$\pm$0.02} & \textbf{0.07$\pm$0.00} & \textbf{0.18$\pm$0.01} \\
                     & \multicolumn{1}{l}{}   & \multicolumn{1}{l}{}   & \multicolumn{1}{l}{}   \\
                     & \multicolumn{1}{l}{}   & \multicolumn{1}{l}{}   & \multicolumn{1}{l}{}   \\ \hline
\end{longtable}

\textbf{MAPE}:
Similar to MAE,  we observe significant improvements in the MAPE values across all three base models. For \citet{Duffy2022}, MAPE improves by 0.09 percentage points~(pp) in IVS, by 0.04 pp in LVID, and by 0.05 pp in LVPW. In the case of \citet{Gilbert2019}, MAPE decreases by 0.04 pp in IVS, by 0.04 pp in LVID, and by 0.05 pp in LVPW. Similarly, for \citet{Wan2023}, MAPE improves by 0.07 pp in IVS, by 0.04 pp in LVID, and by 0.10 pp in LVPW. These results demonstrate that the AMM space based alignment consistently leads to improvements in MAPE.

\section{Comparison using Success Detection Rate~(SDR)}\label{sec:sdr}
\begin{figure*}
	\centering
    \includegraphics[scale=0.15]{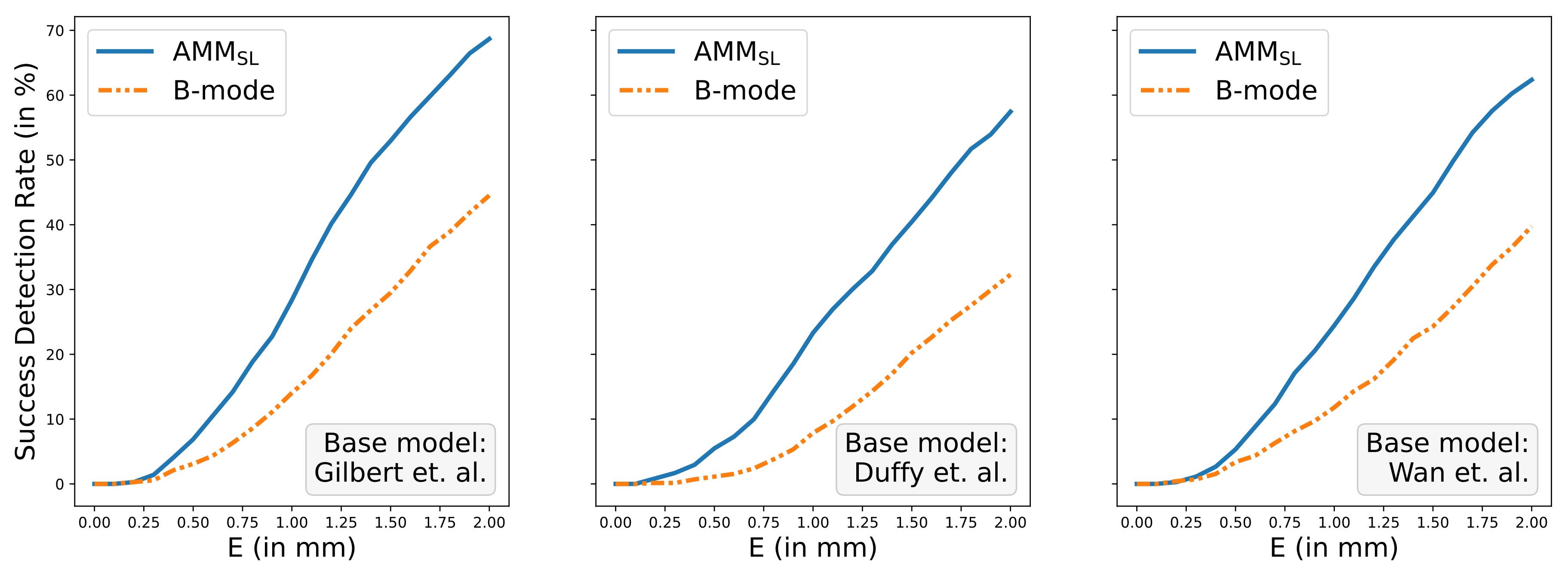}
	\caption{Success Detection Rate~(SDR) for B-mode  and AMM$_{\text{SL}}$ Training for different values of error tolerance E, ranging from 0-2 mm. }
	\label{fig:delta_curve}
\end{figure*}
Comparative metrics are introduced to evaluate the success rates of two training modes within a specified error threshold of $E$ mm. The success detection rate (SDR) is defined as:  
\[
SDR = \frac{1}{|M_{\text{test}}|} \sum_{m=1}^{|M_{\text{test}}|} \mathds{1}_m[\text{MAE}_{\text{Avg}} \leq E]
\]  
Here, $\text{MAE}_{\text{Avg}}$ is calculated as the average mean absolute error (MAE) across all structures: IVS, LVID, and LVPW, for each sample $m$ in the test set ($M_{\text{test}}$). Figure~\ref{fig:delta_curve} presents SDR plots across error tolerances \( E \) ranging from 0 to 6 mm for the three different base models. 
The evaluation of AMM training versus B-mode training reveals distinct improvements in success detection rates for LV linear measurements, particularly within error tolerances ranging from 0 to 2 mm, which are critical for clinical diagnoses. For all base models, an improvement of approximately 15 percentage points~(pp) is observed for an error tolerance of 1 mm, while around 20 pp improvement is seen for an error tolerance of 2 mm.

\section{Comparision with human expert}
\begin{figure*}[t!]
\centering
	\includegraphics[scale=0.27]{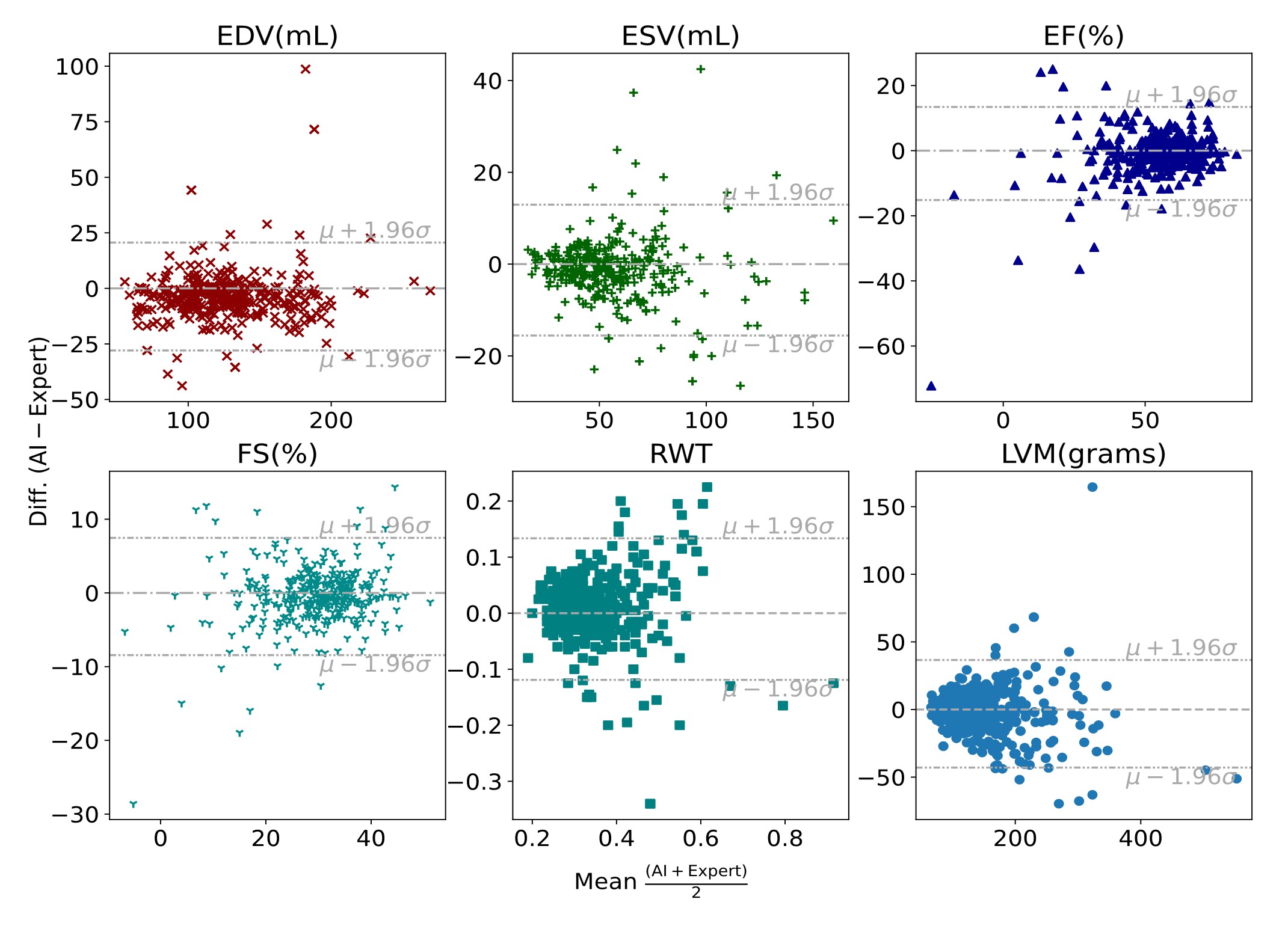}
	\caption{Bland-Altman~\citep{bland-altman} plots comparing cardiac function evaluations derived from human expert annotations and predictions from the AMM-trained base model~(\citet{Duffy2022}).}
	\label{fig:cardiac_function}
\end{figure*}
\subsection{Cardiac Function Evaluation}\label{subsec:blandaltman}
In this section, we compare various cardiac function predictions made by expert annotators with the AMM$_{\text{SL}}$ baseline. \textit{Left Ventricular Mass (LVM)} plays a crucial role in assessing the presence of cardiac hypertrophy and understanding the structural adaptations of the heart in response to conditions such as hypertension or aortic stenosis. Accurate estimation of \textit{LVM} with AMM$_{\text{SL}}$ significantly improves the ability to stratify patient risk and monitor the progression or regression of hypertrophy, which is essential for making informed therapeutic decisions. \textit{Relative Wall Thickness (RWT)} quantifies the ratio of the left ventricular wall thickness to the chamber dimensions, serving as an important indicator of concentric remodeling and hypertrophy. With \textit{RWT} measurements, enhanced precision leads to more accurate classification between concentric and eccentric hypertrophy, reducing diagnostic ambiguities and enabling better-targeted interventions. \textit{Fractional Shortening (FS)}, an alternative method for evaluating systolic function, provides an estimate of the percentage change in left ventricular diameter during systole. \textit{FS} measurement benefits from AMM$_{\text{SL}}$'s ability to reduce variability in diameter measurements, thus improving the accuracy of systolic function evaluation, especially in cases where volume-based measures are less reliable. \textit{End-Diastolic Volume (EDV)} and \textit{End-Systolic Volume (ESV)} are critical measurements for assessing both diastolic and systolic function of the left ventricle, respectively. The precision achieved through AMM$_{\text{SL}}$ reduces measurement variability, enabling more reliable assessments of cardiac filling and contractility. Finally, \textit{Ejection Fraction (EF)}, derived from both \textit{EDV} and \textit{ESV}, is a key marker for diagnosing heart failure. Enhanced accuracy in \textit{EF} measurement not only improves the ability to detect subtle declines in systolic performance but also aids in better decision-making for patient management and treatment planning.

\subsection{Correlation Plot}
\label{subsec:correlation}
\begin{figure*}[t!]
\centering
	\includegraphics[scale=0.2]{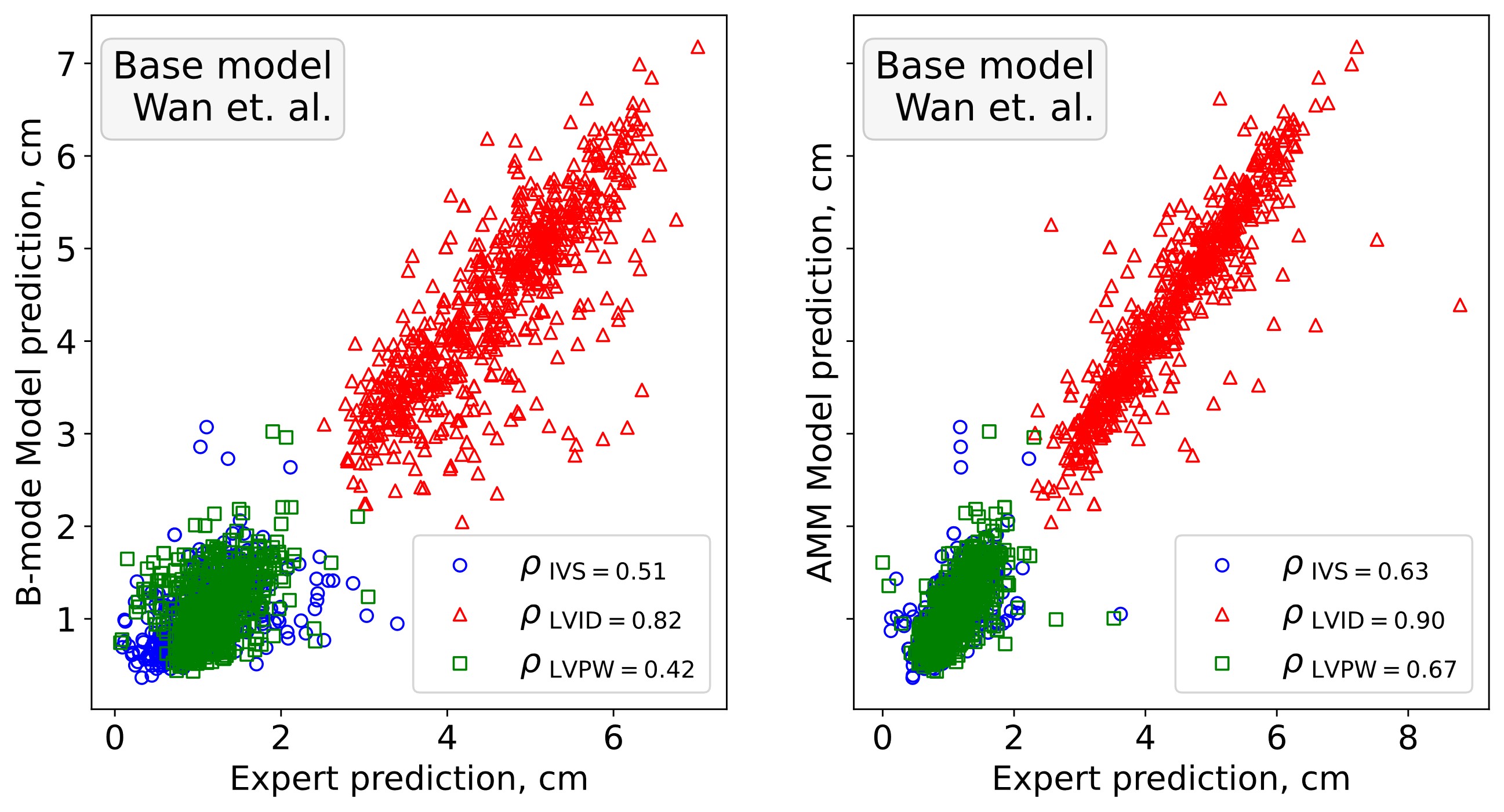}
	\caption{Correlation between base model~(\citet{Wan2023}) predictions from the B-mode training and AMM-training with expert annotation for different LV structures}
	\label{fig:correlation}
\end{figure*}

\begin{figure*}[t!]
\centering
	\includegraphics[scale=0.3]{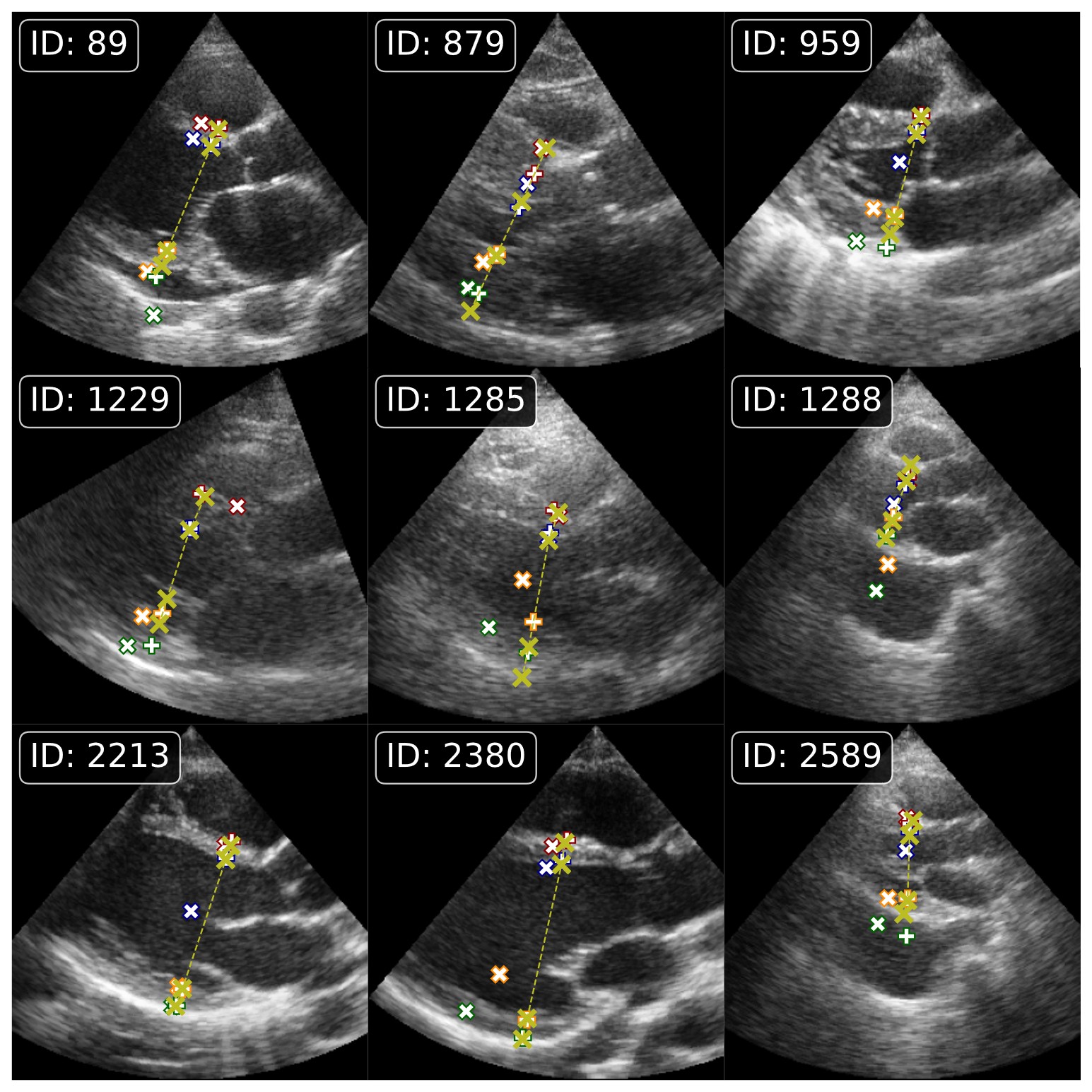}
	\caption{Visualization with base model~(\citet{Gilbert2019}) predictions in cases where AMM-based training reduces B-mode errors by aligning the landmarks along a user-provided scanline.}
	\label{fig:visi:1}
\end{figure*}

Figure~\ref{fig:correlation} illustrates the Pearson correlation coefficients~($\rho$) for B-Mode and AMM$_{\text{SL}}$ length measurements for the base model ~\citet{Wan2023}, derived from predicted landmarks and compared with expert annotations across the test-split data for all LV structures—IVS, LVID, and LVPW. For the B-Mode training, the correlation coefficients are $\rho_{\text{IVS}} = 0.51$, $\rho_{\text{LVID}} = 0.82$, and $\rho_{\text{LVPW}} = 0.42$. In comparison, the AMM$_{\text{SL}}$ model shows notable improvements in the correlation coefficients for $p_{\text{IVS}}$ and $p_{\text{LVPW}}$, which increase to 0.61 and 0.68, respectively, when compared to human annotations. Additionally, $p_{\text{LVID}}$ improves from 0.82 to 0.90. These improvements can be primarily attributed to the AMM's constraint of predicting landmarks along a straight line, which minimizes the misalignment of landmarks and reduces the errors associated with over- or under-estimation in the measurements. This demonstrates that AMM$_{\text{SL}}$ effectively mimics the performance of human experts in landmark-based length measurements.

\begin{figure*}[t!]
\centering
	\includegraphics[scale=0.20]{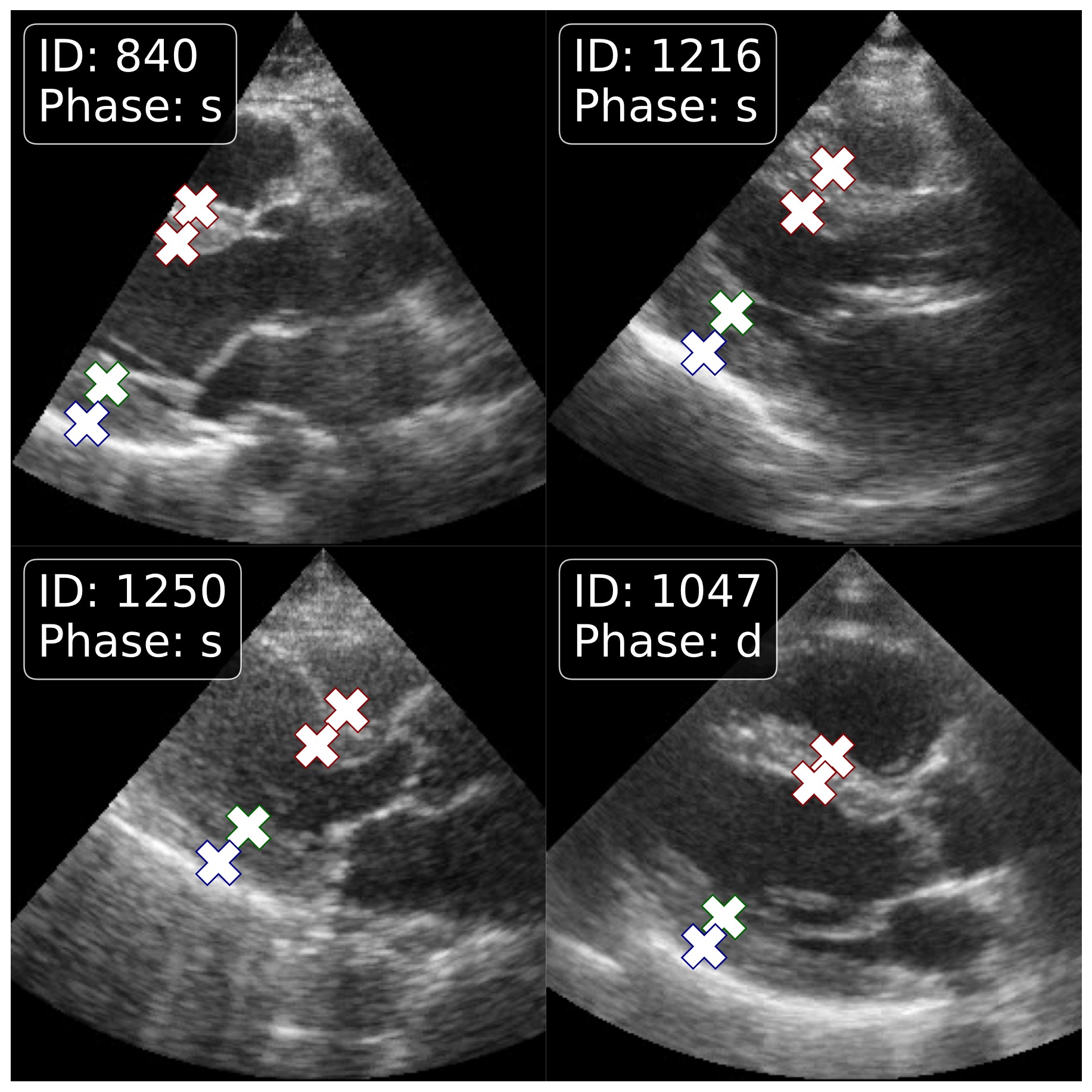}
	\caption{Visualization with base model~(\citet{Gilbert2019}) predictions for the cases where septal bulge is present}
	\label{fig:visi:2}
\end{figure*}

\section{Applications of Our Proposed Approach: EnLVAM}\label{sec:visualization}
\subsection{Addressing Over/Under estimation in B-Mode approaches}
\label{subsec:applications}
As depicted in Figure~\ref{fig:visi:1}, restricting landmark prediction to the SL during AMM training results in a notable improvement in CE across all scenarios, ultimately enhancing the performance of automatic measurements. This improvement is attributed to the elimination of the ambiguous landmark predictions often observed in B-mode training, where predicted landmarks are prone to shifting along the myocardial wall. Such shifts can lead to inaccuracies in LV dimension estimations, resulting in over- or underestimation. By focusing predictions on the SL, this approach ensures greater precision in landmark placement, thereby reducing errors and improving the clinical relevance of LV linear measurements.

\subsection{Conducting Measurements for Septal Bulge Cases}

In addition to addressing landmark prediction shifts, the AMM training methodology incorporates flexibility to accommodate anatomical variations, such as septal bulge. As shown in Figure~\ref{fig:visi:2}, this flexibility allows accurate measurements to be performed even in cases where septal morphology deviates from typical patterns. This adaptability not only enhances the robustness of the AMM technique but also ensures that clinically meaningful measurements can be obtained across a wider range of patient presentations, further underscoring the value of this approach in clinical practice.

\section{Conclusion}\label{sec:conclusion}
This work addresses the challenges associated with accurately estimating LV linear measurement in prior B-mode approaches. Those methods depend on heatmap-based landmark detections, which can create ambiguity in the placement of landmarks and lead to inaccurate measurements potentially resulting in the misdiagnosis of cardiovascular disease. Additionally, various medical studies recommend performing LV linear measurements at either the mitral leaflet tip level or mid-ventricular level, leading clinicians to have individualized preferences for each patient. However, previous B-mode approaches have been limited to one of the recommended locations, therefore making it difficult to accommodate these preferences. To address this challenge, we propose a novel semi-automatic method utilizing AMM imaging derived from virtual SL to perform LV landmark predictions on AMM images.
Our approach offers two key benefits: it aligns landmark predictions with the user-provided scan line in B-mode, and it gives users the flexibility to perform LV linear measurements at different locations.  By combining the advantages of AMM imaging with a user-centric approach to scan line alignment, our method represents a step forward in improving the precision and reliability of LV linear measurement. 

In the future, our work can be combined with automatic approaches that propose the positioning of the virtual SL. This allows for further automation of the process and would limit the manual intervention of the clinicians to corrections of the SL. Additionally, an interesting avenue of future work is to combine B-mode predictions with the AMM model to provide more robust multiview LV linear measurements.

\end{document}